\documentclass[conference]{IEEEtran}

\ifCLASSOPTIONcompsoc
  \usepackage[nocompress]{cite}
\else
  \usepackage{cite}
\fi
\ifCLASSINFOpdf
\usepackage[pdftex]{graphicx}
\else
\fi
\usepackage{array,multirow}
\usepackage[float]{}

\usepackage{algorithm}
\usepackage{algpseudocode}
\usepackage{textcomp}
\usepackage[english]{babel}
\usepackage[utf8x]{inputenc}
\usepackage{amsmath}
\usepackage{amssymb}
\usepackage{graphicx}
\usepackage{subcaption}
\usepackage[export]{adjustbox}

\usepackage{cleveref}

\usepackage[colorinlistoftodos]{todonotes}
\graphicspath{ {} }
\makeatletter
\DeclareTextCommandDefault{\textleftarrow}{\mbox{$\m@th\leftarrow$}}
\makeatother

\usepackage{fancyhdr}

\begin{document}
%
\title{General Video Game AI:\\ Learning from Screen Capture}
%
%
%
%

\author{\IEEEauthorblockN{Kamolwan Kunanusont}
\IEEEauthorblockA{University of Essex\\
Colchester, UK\\
Email: kkunan@essex.ac.uk}
\and
\IEEEauthorblockN{Simon M. Lucas}
\IEEEauthorblockA{University of Essex\\
Colchester, UK\\
Email: sml@essex.ac.uk}
\and
\IEEEauthorblockN{Diego P\'erez-Li\'ebana}
\IEEEauthorblockA{University of Essex\\
Colchester, UK\\
Email: dperez@essex.ac.uk}}





\IEEEtitleabstractindextext{%
 \thispagestyle{plain}
          \fancypagestyle{plain}{
            \fancyhf{} 
            \fancyfoot[L]{978-1-5090-4601-0/17/\$31.00~\copyright2017~IEEE} 
            \renewcommand{\headrulewidth}{0pt}
            \renewcommand{\footrulewidth}{0pt}
          }
\begin{abstract}
General Video Game Artificial Intelligence is a general game playing framework for Artificial General Intelligence research in the video-games domain. In this paper, we propose for the first time a screen capture learning agent for General Video Game AI framework. A Deep Q-Network  algorithm was applied and improved to develop an agent capable of learning to play different games in the framework. After testing this algorithm using various games of different categories and difficulty levels, the results suggest that our proposed screen capture learning agent has the potential to learn many different games using only a single learning algorithm.
\end{abstract}

}

\maketitle
\IEEEdisplaynontitleabstractindextext

%

\ifCLASSOPTIONcompsoc
\IEEEraisesectionheading{\section{Introduction: AGI in Games}\label{sec:introduction}}
\else
\section{Introduction: AGI in Games}
\label{sec:introduction}
\fi

%
%
%
%
\IEEEPARstart{T}{he} main objective of Artificial Intelligence is to develop automated agents that can solve real world problems at the same level as humans. These agents can be divided into two broad types: domain-specific agents and general agents. Domain-specific agents focus on solving only one problem, or a few
, problems at the same or better level than skilled or trained humans. 
For video games, there have been several competitions that encouraged the development of AI players for specific games. Examples are the Ms PacMan \cite{lucas2007ms} competition, which took place between $2008$ and $2011$, and the Mario AI competition \cite{togelius20102009}, held between $2009$ and $2012$. 

However, apart from being excellent at performing a single skilled task, humans are also capable of solving several different types of problems efficiently. For example, in video game terms, humans do not restrict their ability to be expert at only one or a few types of video games, as opposed to domain-specific AI agents. This inspired researchers to study another type of AI agent called ``general'' agents. The word \textit{general} in this context means that the intelligence embedded in such agents should be applicable for many types of problems. This is not necessarily equivalent to combining different algorithms from domain-specific agents to create a general agent, but to develop only one that is general enough to adapt with all tasks, in an Artificial General Intelligence (AGI) setting~\cite{goertzel2007artificial}.


To efficiently evaluate the generality of an AGI agent, AGI framework tasks should not be finite, but updated frequently to ensure that developed AGIs are not domain-specific with the seen problems. General Video Game Artificial Intelligence or GVG-AI \cite{perez20152014} is a General Video Game Playing \cite{levine2013general} framework with this characteristic. Video Game Description Language was applied \cite{schaul2013video} to easily design and develop new video games, increasing the number of games from 30 games (when it was first started in 2014) to 140 games at the time this paper is written (January 2017).

Human video game players receive most of their information through visual sensors (e.g. eyes), and interact by giving actions directly via game controllers. This inspired attempts to develop video game automated players using mainly screen information as an input, such as the PacMan screen capture competition \cite{competition_2007} and VizDoom \cite{kempka2016vizdoom}. 
An important breakthrough of this is the Deep Q-Network, proposed by Mnih et al.~\cite{mnih2015human}. The developed agent was evaluated using the Arcade Learning Environment (ALE) \cite{bellemare2012arcade} framework. Since the algorithm receives screen information as input and produces actions as output, it is adaptable to many different domains.
This paper presents a work that applies a Deep Q-Network to the GVG-AI framework, in order to develop a screen capture learning agent in this framework for the first time, as far as the authors are aware
. As previously suggested, ALE's game set is finite, but GVG-AI is not. The purpose of this work is to present another version of a Deep Q-Network for the GVG-AI framework. 

The paper is structured as follows: Section~\ref{sec:review} reviews the related work that has been done, while the relevant background details are described in Section~\ref{sec:background}. This is followed by the proposed learning agent algorithm (Section~\ref{sec:method}) and the experiment results (Section~\ref{sec:exp}), and finally the conclusions and possible future works are discussed in Section~\ref{sec:conc}.

\section{Related Work} \label{sec:review}

The first attempt to apply AGI within the game domain is General Game Playing (GGP) \cite{genesereth2014general}, which is a platform for Artificial General Intelligence for games. Later, ALE was proposed in 2013 by Bellemere et al.\cite{bellemare2012arcade} as a framework to evaluate Artificial General Intelligence, using some of the Atari 2600 video games as tasks to solve. In the same year, General Video Game Playing (GVGP) was defined by extending from GGP \cite{levine2013general}. Unlike GGP, GVGP focuses more on general agents for video games, which require more player-environment real time interactions. Based on GVGP, game information should be encapsulated and given to the player during the game play, allowing some (small) time for the player to determine the next action based on the given information. The first GVGP competition and framework is General Video Game Artificial Intelligence or GVG-AI \cite{perez20152014}.

Since the GVG-AI competition first started in 2014, there have been several works aimed at developing GVGP agents, although most of them are based on planning algorithm 
due to framework restrictions (i.e. no replay and timing constraints). The most popular algorithm applied was Monte Carlo Tree Search (MCTS) \cite{browne2012survey}. 
There have been attempts to modify MCTS to work more efficiently with GVG-AI, such as using evolutionary algorithm 
with knowledge-based fitness function to guide rollouts \cite{perez2014knowledge}, or storing statistical information in tree nodes instead of pure state details \cite{perez2015open}.\par
There was a claim that GVG-AI will operate a new track called learning track to encourage learning agent development in near future \cite{perez2016general}
. Only one learning agent has been proposed so far, based on neuro-evolution \cite{samothrakis2015neuroevolution}. The framework was adjusted so the agent could replay games, and the forward model was made inaccessible
. Based on this, the learning agent was obliged to rely only on its own gameplay state observation and experience. We employed similar framework adjustments in this work
, where only the level map dimension, block size and screen information are accessible to our learning agent.

Learning from visual information has been taken into account for years, and some video game research frameworks such as the Ms. PacMan screen capture competition \cite{competition_2007} 
and ALE, have screen capture tools embedded. 
Image recognition algorithms can be used to obtain user-specified features. Also, recently flourish attention in Deep learning \cite{lecun2015deep}, especially a spatial-based deep neural network called Convolutional Neural Network (CNN) \cite{2_nielsen_2016}
, encouraged more adaptations of CNN usages in auto image feature extraction
. After the features are extracted from asynchronous series 
of screens captured during gameplay, Reinforcement learning \cite{sutton1998reinforcement} is usually applied as the learning algorithm. The first framework idea that combines deep learning and reinforcement learning in a visual learning task was proposed by Lange and Riedmiller \cite{lange2010deep}. Later, Mnih et. al. \cite{mnih2015human} proposed a breakthrough learning general agent for ALE. The algorithm they proposed is called Deep Q-Network, which is a combination of a Deep convolution neural network and Q-learning in reinforcement learning. 

To the best of our knowledge, there is no GVG-AI learning agent that uses visual information as an input. A Master thesis done by B. Ross \cite{ross2014general} applies sprite location information in grid observation to guide MCTS towards a new sprite type that never explored 
, although it is still a planning agent. Our paper is the first attempt to develop a screen capture learning agent for GVG-AI.

\section{Background} \label{sec:background}

\subsection{Convolutional neural network}
\begin{figure}[!t]
    \centering
        \includegraphics[width=\columnwidth*4/5]{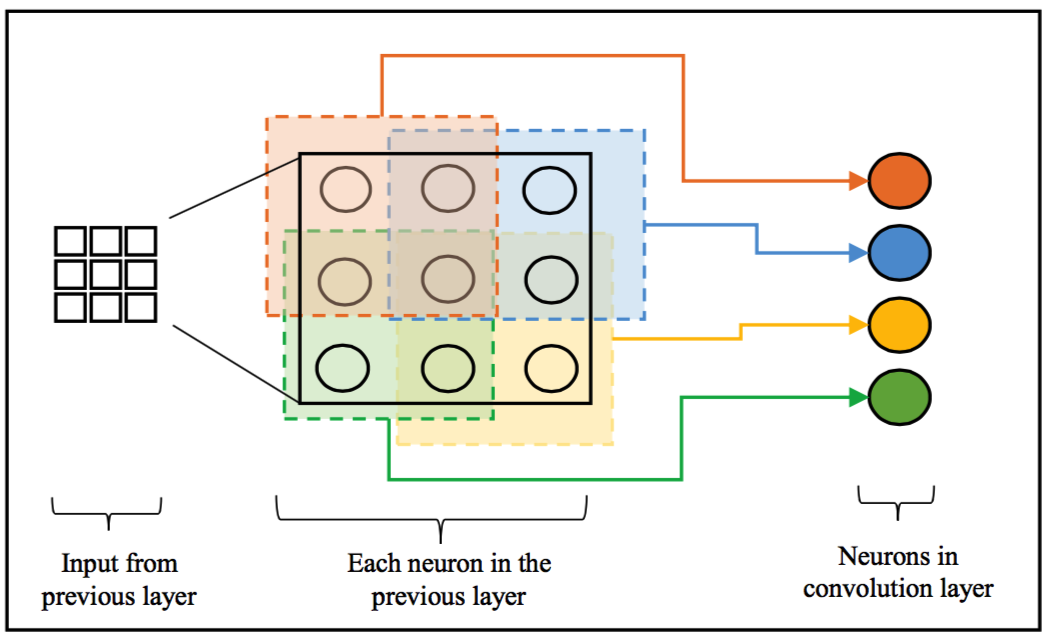}
          \caption{An example of convolution layer}
          \label{fig:clayer} 
\end{figure}

Convolutional neural networks are a type of neural network that was designed for image-like data feature extraction. The concept was first introduced in 1998 \cite{lecun1998gradient}, but gained more interest after being successfully applied as a part of classifier algorithm for ImageNet \cite{krizhevsky2012imagenet}. For each convolution layer, each neuron is responsible for one value of input data (i.e. a pixel of RGB value for image input). The idea is that images in the same category often share the same features in certain local areas: for example, the pictures of dogs are most likely to contain dog ears at some location. The image pixels that represent dog ears share the same or similar features, even though they are not located at the same locations in the images. Convolution layers extract this by passing data from the same neighborhood areas into the same neurons, as illustrated in Figure \ref{fig:clayer}. Each area block dimension is called 'kernel size' and the gap between two blocks is called stride size. in Figure \ref{fig:clayer}, kernel size = $2 \times $2 and stride size = $1 \times $1. In each convolution layer, each output neuron is embedded with a non-linear rectifier function, which in this work is $
f(x) = max(0,x)
$

\subsection{Q-learning}
Q-learning is an off-policy temporal difference learning algorithm in Reinforcement Learning \cite{sutton1998reinforcement} (RL), which is a Machine Learning paradigm that aims to find the best policy to react in the problem it is solving. This is done by maximising reward signals given from the environment for each available action in a given situation. 
Q-learning is a model-free RL technique that allows online learning, and it updates the current state values according to maximum return values of all states after providing available actions. 

A main challenge in Reinforcement Learning is to balance exploitation and exploration. Exploitation chooses the best action found so far, whilst exploration selects another alternative option to improve the current policy. A simple solution for this is to apply an $\epsilon$-greedy policy, which selects an action at random with probability $\epsilon$
. In our implementation, $\epsilon$ was initialised at $1$ and decreased by $0.1$ in every time step until it stabilizes at $0.1$
. 


\subsection{Deep Q-Network}
Deep Q-Network consists of two components: a deep convolution neural network and Q-learning. A CNN is responsible for extracting features for the images and determining the best action to take. The Q-learning component takes these extracted features and evaluates state-action values of each frame.

\subsubsection{Network structure}
The Deep CNN proposed by Mnih et. al. \cite{mnih2015human} consists of 6 layers: 1 input layer, 4 hidden layers and 1 output layer, connected sequentially. The input layer receives pre-processed screenshots and passes them to the first hidden layer. The first three hidden layers are convolutional layers and the last one is a fully-connected dense layer. The output layer has the same number of neurons as the number of ALE possible actions.

\subsubsection{Method}
The method has two main units: pre-processing and training. Pre-processing transforms a $210 \times 160$ pixel screen captured image into an $84 \times 84$ pixel of Y-channel value in RGB. This means the input sizes are fixed for all ALE games tested, unlike GVG-AI games that have different screen sizes. Training is done by passing 4 pre-processed recent frames into the network, giving an action returned by the network to the game and observing the next screen and the reward signal. Input screens, actions performed, result screen and reward (clipped to either -1, 0, 1, based on either positive, unchanged or negative score) are all packed together into an object called ``experience'', which will be stored into the experience pool. Then, some experiences in this pool are sampled to calculate action-value outputs of these four recent frames (of the selected experience). After that, the output is used to update the network connection weights using gradient descent. To randomly select stored experiences and updates, the network based on them is called ``experience replay'' 
which is the core component of this learning algorithm. Also, to achieve a more stable training, the network is cloned and the clone is updated from Q-learning while the original is used to play games. After a while, the original network is reset to the updated clone. In our GVG-AI agent we used experience replay but not double network learning, as the memory usage was not affordable.
\subsection{GVG-AI}
\subsubsection{GVG-AI framework}
The GVG-AI framework contains $140$ games in total, $100$ of which are single player games and the rest, 2-player games. In this paper the developed agent was tested with some of the single player games alone. Video games in this framework were all implemented using a java port of Py-vgdl, which is developed by T. Schaul \cite{schaul2013video}. All game components, including avatars and physical objects, are located within $2$ dimensional rectangular frames. 

The 2014-2016 GVG-AI competition featured only the ‘planning track', which is subdivided into single player and 2-player settings. Submitted agents are not allowed to replay games, instead a forward model is given for future state simulation. All state information is encapsulated into this model and the agent can select an action, observing results before performing the action in the game. However, since a game state space is very large and the agent is allowed up to 40 ms to return an action, exhaustive search is not practical. In this paper, only screen information, screen block size, and level dimension were used as inputs, and all time limit restrictions were disabled. 

\subsubsection{Games tested}
There are 6 single player games used in the experiments described in this paper. These can be categorised into two groups: simple exit-finding games and stochastic shooting games. \textit{Grid-world}, \textit{Escape} and \textit{Labyrinth} are exit-finding games, while \textit{Aliens}, \textit{Sheriff} and \textit{Eggomania} are shooting games. We selected these games specifically because of their similar nature and varying levels of difficulty. For example, \textit{Labyrinth} is more difficult to solve than \textit{Escape}, which is in turn harder than \textit{Grid-world}. Similarly, \textit{Sheriff} is harder than \textit{Aliens} but less difficult than \textit{Eggomania}. Unfortunately, the long time needed to train the networks prevented us from testing on more games. Details of each game can be seen in Table \ref{gameTable}.
\begin{table*}[!t]
\renewcommand{\arraystretch}{3}
\caption{Descriptions of Games Tested}
\label{gameTable}
\centering
\setlength\tabcolsep{3.5pt}
\renewcommand{\arraystretch}{1.2}
\begin{tabular}{|c|c|c|c|c|}
\hline
\bf{Game type} & 
\bf{Score system} & 
\bf{Game name} & 
\bf{\begin{tabular}{@{}c@{}}Winning/Losing \\ condition\end{tabular}} 
& {\bf{\begin{tabular}{@{}c@{}}Number of different \\ sprite type\end{tabular}}} \\
\hline
\multirow{3}{*}{Exit-finding} 
& \multirow{3}{*}{\begin{tabular}{@{}c@{}}Once at the end \end{tabular}}
& Grid-world & \multirow{5}{*}{\begin{tabular}{@{}c@{}}Win: exit reached \\
Lose: timeout or falling into traps  \end{tabular}} & 4 \\ \cline{3-3}\cline{5-5}

& & Escape &  & 5 \\ \cline{3-3} \cline{5-5}
& & Labyrinth & & 4 \\ \cline{1-3} \cline{5-5}

\begin{tabular}{@{}c@{}}Exit-finding with \\ collectable items \end{tabular} &\multirow{6}{*}{\begin{tabular}{@{}c@{}}Accumulative\\during gameplay  \end{tabular}}
& \begin{tabular}{@{}c@{}}Modified \\ Labyrinth\end{tabular} & & 5 \\
\cline{1-1} \cline{3-5}
\multirow{3}{*}{Shooting} 
& 
& Aliens & \begin{tabular}{@{}c@{}}Win: All enemies shot \\ Lose: Hit by a bomb, touched by an enemy \end{tabular} 
& 6
\\
\cline{3-5}
& 
& Sheriff 
& \begin{tabular}{@{}c@{}}Win: All enemies shot \\ Lose: Shot by an enemy, timeout \end{tabular} 
& 7
\\
\cline{3-5}
& 
& Eggomania & \begin{tabular}{@{}c@{}}Win: Enemy shot \\ Lose: Failed to collect one item, timeout \end{tabular}
& 8
\\
\hline
\end{tabular}
\end{table*}

\subsubsection{MCTS}
Monte Carlo Tree Search (MCTS;~\cite{browne2012survey}) is a tree search technique that builds an asymmetric tree in memory, biased towards the most promising parts of the search space, via sampling available actions. This is the best sample controller provided with the GVG-AI framework (\textit{SampleOLMCTS}) and, despite being a planning algorithm, has been chosen in this study to compare  with the performance of our proposed learning agent (in the absence of an actual learning algorithm to compare it with).

\section{Proposed Method} \label{sec:method}

We present a GVG-AI screen capture learning agent based on a Deep Q-Network.
We modified the framework to allow replaying for any created agents, and another alteration made was to disable all timing limits, which include 1 second for constructing an agent and up to 40 ms action determination for each time step. Therefore, our agent is allowed unlimited time in initialisation and learning steps.

\subsection{Deep Q-Network for GVG-AI}
Similar to the original Deep Q-Network, our proposed learning method consists of a pre-processing and a learning unit.
\subsubsection{Pre-processing unit}
\begin{algorithm}[!t]
\caption{Visualize pre-processing (RealPrep)}\label{alg:RealPrep}
\begin{algorithmic}[1]
\small
\State $\text{{Input: Game block size \textit{bSize}}}$
\State $\text{{Output: preprocessed image in 2D array of double format}}$
\State $\text{{BEGIN}}$
\State\hspace{0.5cm} $\text{\textit{Im}  \textleftarrow  \hspace{0.075cm} capture the current screenshot}$
\State\hspace{0.5cm} $\text{{\textit{shrunkIm} \textleftarrow \hspace{0.075cm} Shrink(\textit{Im}, \textit{bSize})}}$
\State\hspace{0.5cm} $\text{{\textit{normShrunk} 
\textleftarrow \hspace{0.075cm} Normalize(\textit{shrunkIm})}}$
\State\hspace{0.5cm} $\text{{\textit{smallestAllowed} \textleftarrow \hspace{0.075cm}	smallest size allowed}}$
\State\hspace{0.5cm} $\text{{\textit{extendedIm}  \textleftarrow \hspace{0.075cm} Extend(\textit{normShrunk}, \textit{smallestAllowed})}}$
\State\hspace{0.5cm} $\text{{RETURN \textit{extendedIm}}}$
\State $\text{{END}}$
\end{algorithmic}
\end{algorithm}

\begin{algorithm}[!t]
\caption{Non-visualize pre-processing (GenPrep)}\label{alg:GenPrep}
\begin{algorithmic}[1]
\small
\State $\text{{Input: Grid observation \textit{grid}, color mapper \textit{Mapper}}}$
\State $\text{{Output: pre-processed image in 2D array of double format}}$
\State $\text{{BEGIN}}$
\State\hspace{0.5cm} $\text{\textit{newImage}  \textleftarrow  \hspace{0.075cm}empty array of same dimension as \textit{grid}}$ 
\State\hspace{0.5cm} $\text{For each sprite type \textit{t} \textleftarrow \hspace{0.075cm} \textit{grid}[\textit{i} \textit{j}]}$
\State\hspace{1.0cm} $\text{{IF \textit{t} was  found before}}$
\State\hspace{1.5cm} $\text{{\textit{Color} \textleftarrow \hspace{0.075cm} \textit{Mapper}[\textit{t}]}}$
\State\hspace{1.0cm} $\text{{ELSE}}$
\State\hspace{1.5cm} $\text{{\textit{Color} \textleftarrow \hspace{0.075cm} random a new color}}$
\State\hspace{1.5cm} $\text{{\textit{Mapper}[\textit{t}]  \textleftarrow \hspace{0.075cm} \textit{Color}}}$
\State\hspace{1.0cm}$\text{{\textit{newImage}[\textit{i}, \textit{j}]  \textleftarrow \hspace{0.075cm} \textit{Color}}}$
\State\hspace{0.5cm} $\text{{\textit{normShrunk} 
\textleftarrow \hspace{0.075cm} Normalize(\textit{newImage})}}$
\State\hspace{0.5cm} $\text{{\textit{smallestAllowed} \textleftarrow \hspace{0.075cm}	smallest size allowed }}$
\State\hspace{0.5cm} $\text{{\textit{extendedIm}  \textleftarrow \hspace{0.075cm} Extend(\textit{normShrunk}, \textit{smallestAllowed})}}$
\State\hspace{0.5cm} $\text{{RETURN \textit{extendedIm}}}$
\State $\text{{END}}$
\end{algorithmic}
\end{algorithm}
Since GVG-AI framework supports both visualise and non-visualise game running
, we proposed two pre-processing algorithms to support both representations. The visualise algorithm directly captures the screen image, shrinking each block down into one pixel, normalising the RGB value and expanding into the smallest size allowed. The reason behind this size-modifying algorithm 
is that the same network structure was applied for every game tested, each of which differ in screen widths and heights. It is possible that the screen is too small for the network, therefore it must be extended into a specific size to prevent errors. Large images do not trigger this problem. Lines $4$-$7$ of Algorithm~\ref{alg:RealPrep} show screen capture, image shrinking and normalising respectively, while image extension steps are shown in lines $8$ and $9$.

Non-visualisation pre-processing generates screen information from a framework-provided object called \textit{gridObservation}, which contains all sprite location information at that state. Each sprite type is mapped into a random RGB colour first, then an image is generated based on the \textit{gridObservation} information. After that, this image is normalised and expanded if necessary. Algorithm \ref{alg:GenPrep} shows how to generate an input image from a grid observation. It begins with an empty 2D array creation, then fills each cell with a stored colour (if the same sprite type in that position was found before (lines $6$ and $7$)), otherwise another color is randomly generated and filled into that cell (lines $8$ to $10$). After that, the 2D array is normalised and expanded, as in lines $12$ to $14$.


\begin{figure}[!t]
    \centering
      \begin{subfigure}{0.43\textwidth}
        \includegraphics[width=\columnwidth]{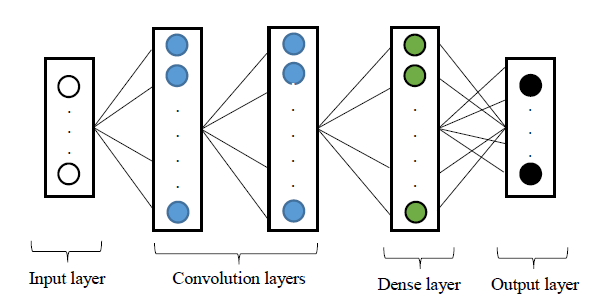}
          \caption{4-layer network}
          \label{fig:4layernet} 
      \end{subfigure}
      \begin{subfigure}{0.43\textwidth}
        \includegraphics[width=\columnwidth]{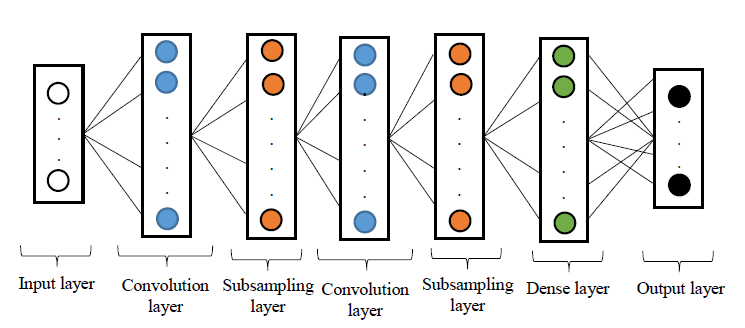}
          \caption{6-layer network}
          \label{fig:6layernet} 
      \end{subfigure}
\caption{Network structure.
\label{fig:mynetwork}
}
\end{figure}
\subsubsection{Learning unit}
A Java deep learning library called DeepLearning4j\footnote{https://deeplearning4j.org/} was applied to create and train the previously designed CNN. Two network structures, as shown in Figure \ref{fig:mynetwork}, were implemented. An input layer consists of $w \times h$ neurons while w and h are width and height of the pre-processed game screen respectively. Convolution layer kernel sizes are either $5 \times 5$ or $3 \times 3$, depending on which network parameter set is chosen. There are $32$ and $64$ neurons in the first and second convolution layers respectively. Stride size is always equal to $1 \times 1$ to capture the most information. Subsampling layers kernel size is $3 \times 3$. Dense layers consist of $512$ neurons fully-connected. Output layer has the same neuron number as available actions of the game. Notice that input and output layer neuron numbers are different for each game but the rest of the network are the same.

Learning procedures for each timestep are summarized in Algorithm \ref{alg:Act}. 
Lines $4$ and $5$ are executed only in visualise mode when the screenshot is taken and pre-processed, while lines $6$ to $8$ are for the non-visualisation mode where the screen input is generated from the grid observation. For the first time step, there are no experiences in the pool, therefore a random action is selected and stored in a newly created experience, along with the current screenshot. Then the action is performed into the game; lines $12$ to $16$ represent these steps. From the second time step onwards, the current screenshot is stored as the result of the previously-created experience, along with the reward signal, as stated in lines $18$ and $19$. This experience is then updated using Q-learning (line 20) before being added into the pool (line 21). Then a new experience is created to store the current screenshot. After that, some experiences are picked from the pool, updated using Q-learning and passed into the network to train. The experience sample and network training steps are given in lines $26$ to $32$. $\epsilon$-Greedy policy is applied to select an action from the network, giving the current screenshot as input. The action is then stored in the experience and given to the network. All of these steps are repeated until the game terminates.

\begin{algorithm}[!t]
\caption{Learning procedures for each timestep}\label{alg:Act}
\begin{algorithmic}[1]
\small
\State $\text{{Input: Game block size \textit{blockSize}, Initial grid observation \textit{grid}}}$
\State $\text{{Output: Action for this game step}}$
\State $\text{{BEGIN}}$
\State\hspace{0.5cm}
$\text{IF graphical user interface allow}$ 
\State\hspace{1.0cm}
$\text{\textit{Image} \textleftarrow \hspace{0.075cm} RealPrep(\textit{blockSize})}$
\State\hspace{0.5cm}
$\text{ELSE}$ 
\State\hspace{1.0cm}
$\text{\textit{mapper} \textleftarrow \hspace{0.075cm} Agent color mapper}$
\State\hspace{1.0cm}
$\text{\textit{Image} \textleftarrow \hspace{0.075cm} GenPrep(\textit{grid}, \textit{mapper})}$
\State\hspace{0.5cm}$\text{expPool \textleftarrow \hspace{0.075cm} experience pool}$
\State\hspace{0.5cm}$\text{exp \textleftarrow \hspace{0.075cm} current experience}$
\State\hspace{0.5cm}$\text{Q \textleftarrow \hspace{0.075cm} current state-action value}$
\State\hspace{0.5cm}$\text{IF first time step}$
\State\hspace{1.0cm}
$\text{\textit{exp}[\textit{previous}] \textleftarrow \hspace{0.075cm}\textit{Image}}$
\State\hspace{1.0cm}
$\text{act \textleftarrow \hspace{0.075cm} a random action}$
\State\hspace{1.0cm}
$\text{\textit{exp}[\textit{action}] \textleftarrow \hspace{0.075cm} act}$
\State\hspace{1.0cm}
$\text{RETURN act}$
\State\hspace{0.5cm}$\text{ELSE}$
\State\hspace{1.0cm}
$\text{\textit{exp}[\textit{result}] \textleftarrow \hspace{0.075cm} \textit{Image}}$
\State\hspace{1.0cm}
$\text{\textit{exp}[\textit{reward}] \textleftarrow \hspace{0.075cm} reward of this state}$
\State\hspace{1.0cm}
$\text{\textit{QLearningUpdate}(\textit{exp}, \textit{Q})}$
\State\hspace{1.0cm}
$\text{Add \textit{exp} to \textit{expPool}}$
\State\hspace{1.0cm}
$\text{\textit{exp} \textleftarrow \hspace{0.075cm} $\varnothing$}$
\State\hspace{1.0cm}
$\text{\textit{model} \textleftarrow \hspace{0.075cm} Agent network model}$
\State\hspace{0.5cm}
$\text{\textit{exp}[\textit{previous}] \textleftarrow \hspace{0.075cm} \textit{Image}}$
\State\hspace{0.5cm}
$\text{\textit{trainData} \textleftarrow \hspace{0.075cm} $\varnothing$}$
\State\hspace{0.5cm}$\text{REPEAT}$
\State\hspace{1.0cm}
$\text{\textit{randExp} \textleftarrow \hspace{0.075cm} pick one experience from \textit{expPool}}$
\State\hspace{1.0cm}$\text{\textit{QLearningUpdate}(\textit{randExp}, \textit{Q})}$
\State\hspace{1.0cm}
$\text{\textit{toTrain} \textleftarrow \hspace{0.075cm} create training data from \textit{randExp} and \textit{Q}}$
\State\hspace{1.0cm}$\text{Add \textit{toTrain} to \textit{trainData}}$
\State\hspace{0.5cm}$\text{UNTIL batch size reaches}$
\State\hspace{0.5cm}$\text{Fit \textit{trainData} to \textit{model}}$
\State\hspace{0.5cm}$\text{With probability 1 - $\epsilon$}$
\State\hspace{1.0cm}$\text{\textit{act} \textleftarrow \hspace{0.075cm} feed \textit{Image} to \textit{model} and get the output action}$
\State\hspace{0.5cm}$\text{ELSE}$
\State\hspace{1.0cm}$\text{\textit{act} \textleftarrow \hspace{0.075cm} randomly select an action}$
\State\hspace{0.5cm}$\text{\textit{exp}[\textit{action}] \textleftarrow \hspace{0.075cm} \textit{act}}$
\State\hspace{0.5cm}$\text{Set current experience to \textit{exp}}$
\State\hspace{0.5cm}$\text{RETURN \textit{act}}$
\State$\text{END}$
\end{algorithmic}
\end{algorithm}
 
Our agent performs a Q-learning update in three occasions during gameplay, in addition to the original DQN that is done only once during experience replay. These include one at the experience creation, one during sampling (as the original DQN) and one more at the end of the episode. This prioritizes the experiences that related to the game results. Also, apart from performing the normal Q-learning equation using a pure score, we added a $-5$ penalty reward to the actions which previous and result screens are the same. This is based on the assumption that the action does not change anything in the game, preventing the agent from getting stuck at the wall and encouraging more movements.

\section{Experiment Results} \label{sec:exp}

\begin{table}[!t]
\scriptsize
\renewcommand{\arraystretch}{3}
\caption{Tuned Parameters}
\label{table:parameter}
\centering
\setlength\tabcolsep{2pt}
\renewcommand{\arraystretch}{1.2}
\begin{tabular}{|c|c|c|}
\hline
\bf{Parameter name} & 
\bf{Value applied} & 
\bf{Meaning}
\\
\hline

batch size & 200, 400 & \begin{tabular}{@{}c@{}}Number of experiences passed \\ in experience replay\end{tabular}
\\
\hline
First kernel size &($5 \times $5), ($3 \times $3) & First convolution layer kernel size
\\ 
\hline
Second kernel size & ($3 \times $3) & Second convolution layer kernel size
\\ 
\hline
Dropout & 0, 0.15, 0.3 & Network dropout value
\\ 
\hline
Subsampling & True, False & With / without subsampling layer
\\ 
\hline
\end{tabular}
\end{table}
 
\subsection{Parameter tuning}
A pre-experimental phase was carried out to select a high quality parameter set. All tuned parameters are described in Table~\ref{table:parameter}. Escape level 0 was selected to do parameter tuning as the nature of the game is simple, the game screen size is not large, and it requires only $16$ moves to win the game optimally. Three criteria were applied in each set for performance measurement. This includes the number of steps taken to win, win percentage at each episode number, and win percentage calculated since the first training. For each criteria, the average, best value and optimal (if applicable) value were measured and compared. We found that, among the selected values, batch size = $400$, kernel size = $5 \times 5$ connected by $3 \times 3$, dropout = $0$ and no subsampling layer gave the best performance in this game. Next, we applied this parameter set to create a CNN, which was embedded into the game player agent to play 6 selected games. The experiment was done 5 times for each game and all results were averaged.

\subsection{Testing with games}
We trained our agent, embedded with a network created with previously tuned parameters, separately for each game. Only the non-visualized pre-processing mode was used because for faster execution. Each experiment was done $5$ times for each game and the results were averaged and compared with the result from MCTS planning-agent, which were averaged from $100$ runs in the original competition-setting framework (with timing constraints and no replay allowed). With different framework settings, the comparison between our learning algorithm and MCTS is not totally 'fair' since MCTS was allowed limited time to think. However, the purpose of this comparison is to measure how well our learning agent played each game compared to the best sample controller.

\subsubsection{Grid-world}
\begin{figure}[!t]
	\hspace{10px}
      \begin{subfigure}{0.4\columnwidth}
        \includegraphics[width=\textwidth,right]{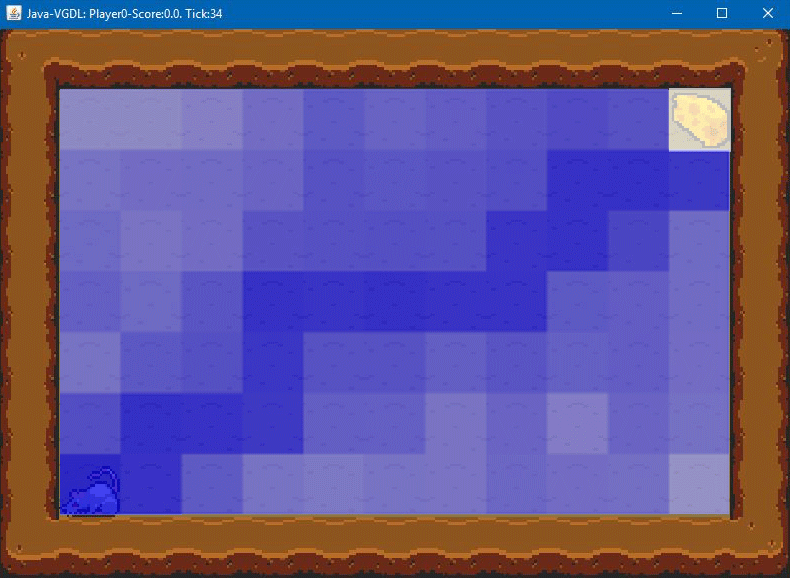}
          \caption{Original Grid-world}
          \label{fig:ori_gw} 
      \end{subfigure}%
      \hspace*{\fill}
      \begin{subfigure}{0.4\columnwidth}
        \includegraphics[width=\textwidth,left]{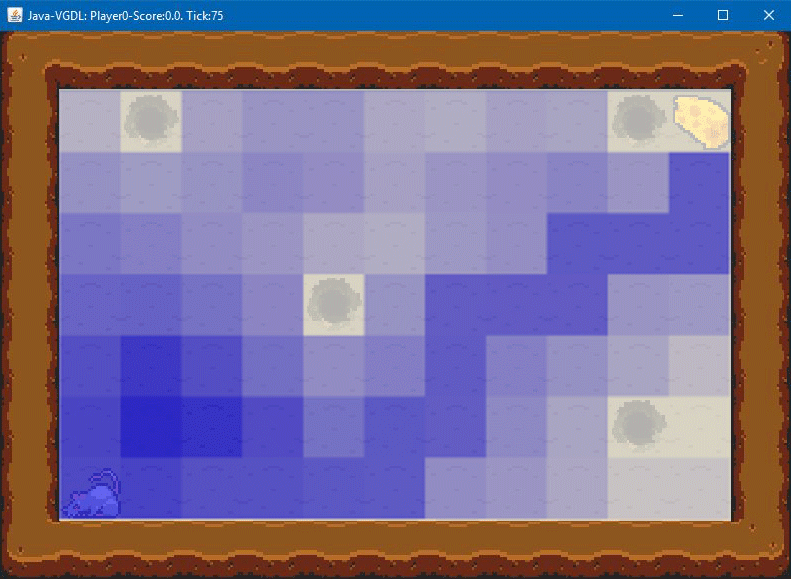}
          \caption{Trapped Grid-world}
          \label{fig:trap_gw} 
      \end{subfigure}
      \hspace{10px}
      \newline  
\caption{Grid-world maps and heat maps.
\label{fig:gw_map}
}
\end{figure}

Original and trapped grid-world were applied in this experiment. Figure \ref{fig:gw_map} shows their differences and heat maps of the agent during training. Heat maps indicate how frequently the agent visited each position during gameplay. In this paper, the number of times the agent visited each cell was stored 
in order to create a heat map. It can be seen that the optimal paths are highlighted with the darkest blue shade, indicating a higher presence of the agent in such paths.

\begin{figure}[!t]
\centering
\includegraphics[width=\textwidth/3]{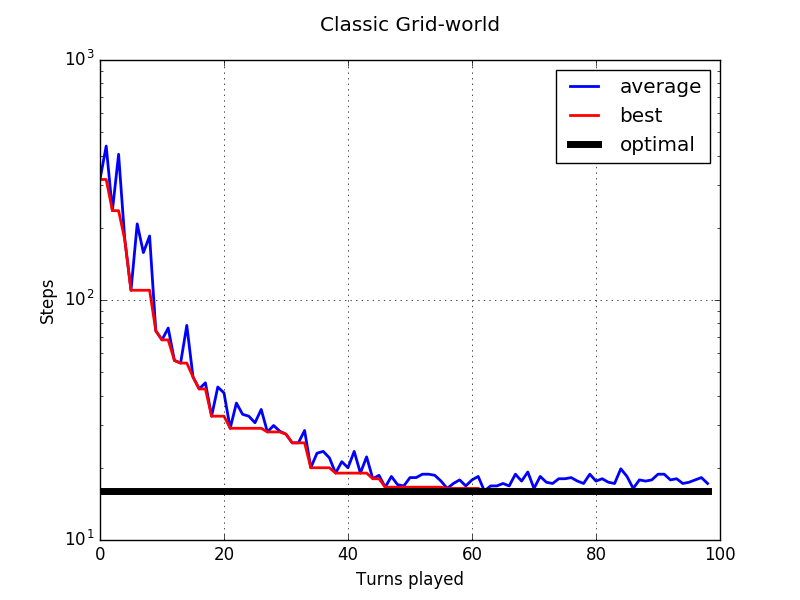}  
\caption{Classic Grid-world steps taken.
\label{fig:ori_gw_steps}
}
\end{figure}

\begin{figure}[!t]
\centering
      \begin{subfigure}{0.35\textwidth}
        \includegraphics[width=\textwidth]{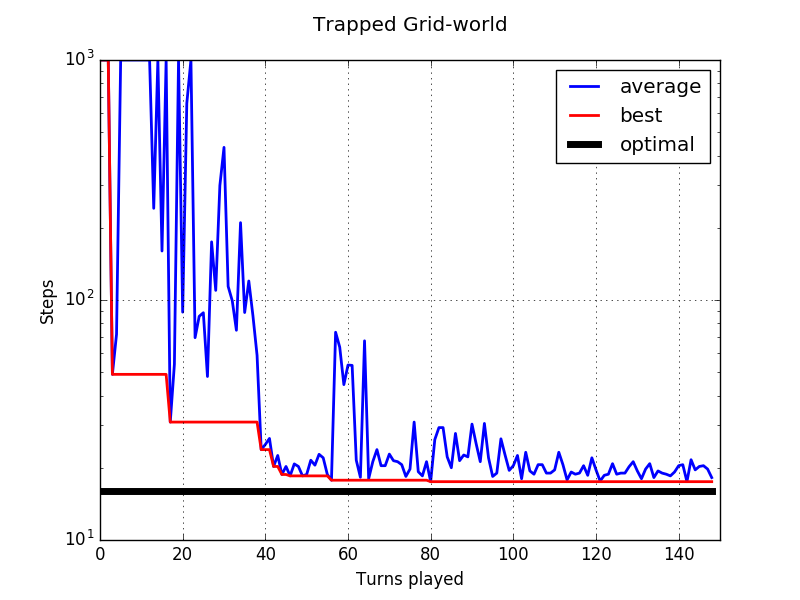}
          \caption{Steps taken}
          \label{fig:trap_gw_steps} 
      \end{subfigure}
       \begin{subfigure}{0.35\textwidth}
        \includegraphics[width=\textwidth]{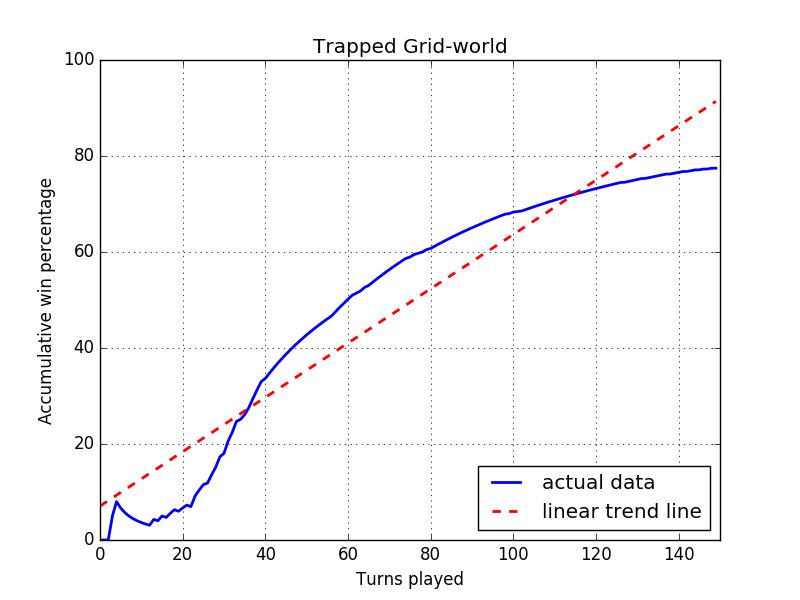}
          \caption{Cumulative win percentage}
          \label{fig:trap_gw_winacc} 
      \end{subfigure} 
\caption{Trapped Grid-world results
\label{fig:gw_trapped}
}
\end{figure}

Figure \ref{fig:ori_gw_steps} shows the steps taken to win for each gameplay for classic Grid-world. It is obvious that the agent took fewer turns to win as it played more, and found the optimal ($16$) steps within $60$ turns. For trapped Grid-world an accumulative win percentage is measured since the agent will lose if the it falls into a hole. The results, as given in Figure \ref{fig:gw_trapped}, show that the agent performance is better when it plays more. This confirms that our agent is capable of learning Grid-world, both original and trapped versions. MCTS agent easily solved Grid-world as it managed to win $79\%$
, although it never found the optimal solution as the minimum steps it reached was $23$ (optimal: $16$). Traps seemed to significantly affect MCTS performance since it never won any in the trapped Grid-world. This suggests that a sequence of actions for trap avoidance could be found from our learning agent but not the MCTS planning agent.

\subsubsection{Escape}
\begin{figure}[!t]
	\hspace{10px}
      \begin{subfigure}{0.4\columnwidth}
        \includegraphics[width=\textwidth]{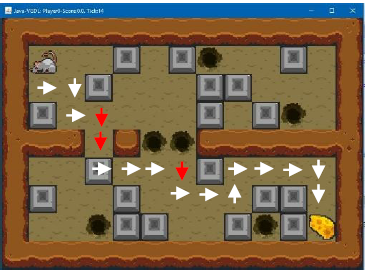}
          \caption{necessary moves}
          \label{fig:escape3_arrow} 
      \end{subfigure}
      \hspace*{\fill}
      \begin{subfigure}{0.4\columnwidth}
        \includegraphics[width=\textwidth]{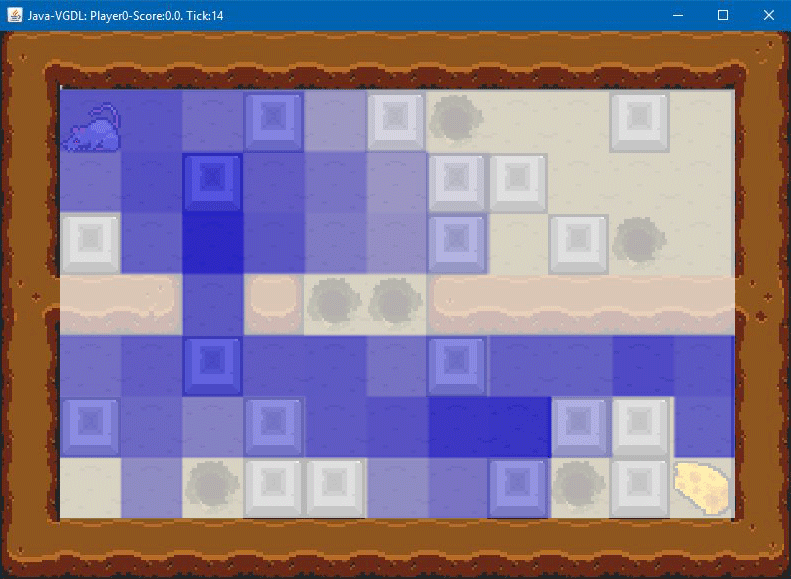}
          \caption{Agent's heat map}
          \label{fig:escape3_hm} 
      \end{subfigure}
      \hspace{10px}
      \newline  
\caption{Escape level 3 necessary moves and heat map
\label{fig:escape3}
}
\end{figure}

\begin{figure}[!t]
\centering
        \includegraphics[width=\textwidth/3]{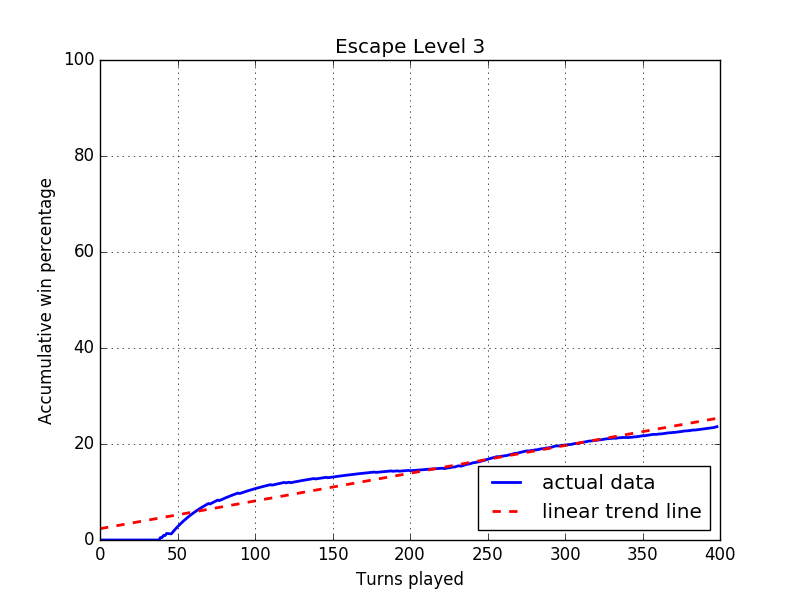}
          \caption{Escape level 3 cumulative win percentage}
\label{fig:escape3_results}
\end{figure}

In addition to parameter tuning, Escape level 3 was selected to test our proposed method. The level 3 game is more challenging than level 0 as it contains two sequences of necessary moves, shown by the red arrows in Figure \ref{fig:escape3_arrow}. Figure \ref{fig:escape3_results} shows cumulative winning percentage for Escape level 3. The necessary moves significantly affected the agent performance as it won at most 3 out of 5 games after playing 350 episodes. However, the trend line of cumulative win percentage increases with more turns played. The heat map in Figure \ref{fig:escape3_hm} shows the darker blue shade trace that leads to the winning position. This suggests that our agent was learning, gradually, to play this game. MCTS agent could not manage to win any single game in Escape at both level 0 and level 3. Again, a sequence of certain actions could be found from learning agent but not planning
.

\subsubsection{Labyrinth}
\begin{figure}[!t]
	\hspace{10px}
      \begin{subfigure}{0.4\columnwidth}
        \includegraphics[width=\textwidth]{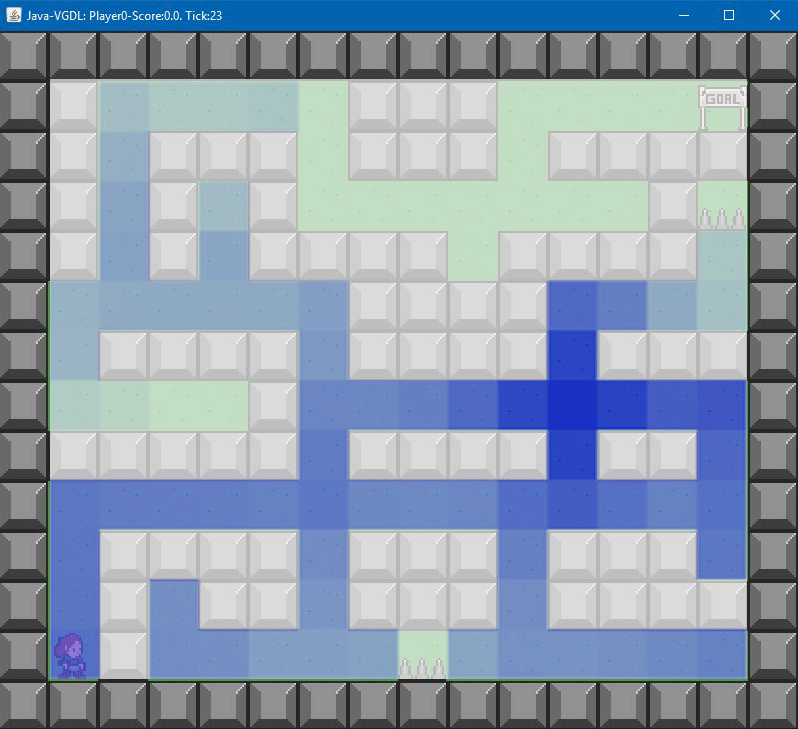}
          \caption{Original}
          \label{fig:laby_ori} 
      \end{subfigure}
      \hspace*{\fill}
      \begin{subfigure}{0.4\columnwidth}
        \includegraphics[width=\textwidth]{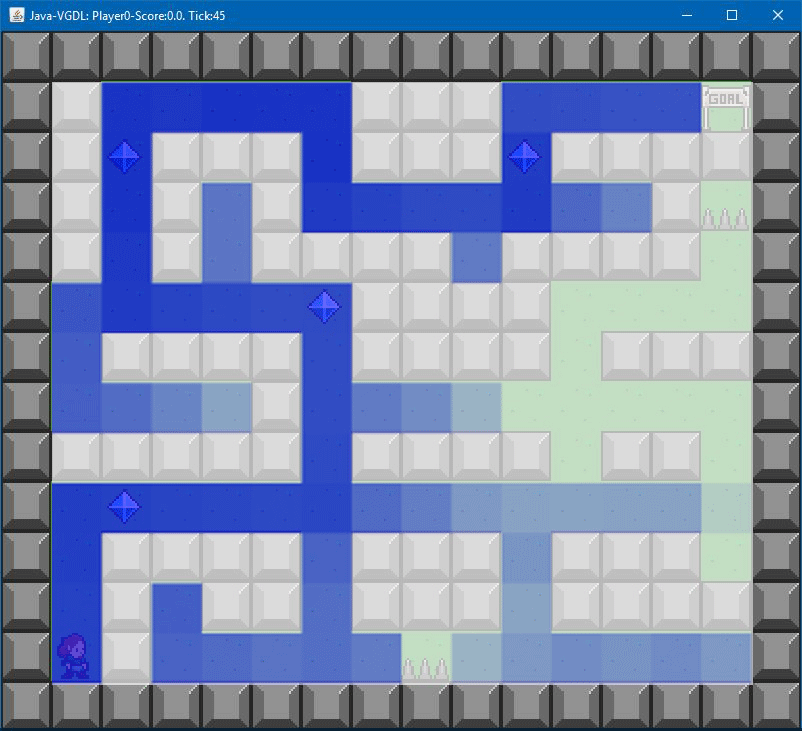}
          \caption{Modified}
          \label{fig:laby_mod} 
      \end{subfigure}
      \hspace{10px}
\caption{Original and Modified Labyrinth heat map
\label{fig:laby_maps}}
\end{figure}

Labyrinth map is much larger than Escape and Grid-world, and contains long corridors in most areas. This means the agent could move only two directions at a time even though four move actions are available, and most of the time the correct moves are sequences of the same actions. Mnih et. at.~\cite{mnih2016asynchronous} mentioned an idea of using immediate reward as a guidance of Reinforcement Learning, which inspired us to add collectible items to Labyrinth. Figure \ref{fig:laby_maps} shows differences between the original and modified Labyrinth, along with the agent’s heat maps. It can be clearly seen that the agent moved into correct directions more with the collectible items that provide reward signals.

\begin{figure}[!t]
\centering
\includegraphics[width=\textwidth/3]{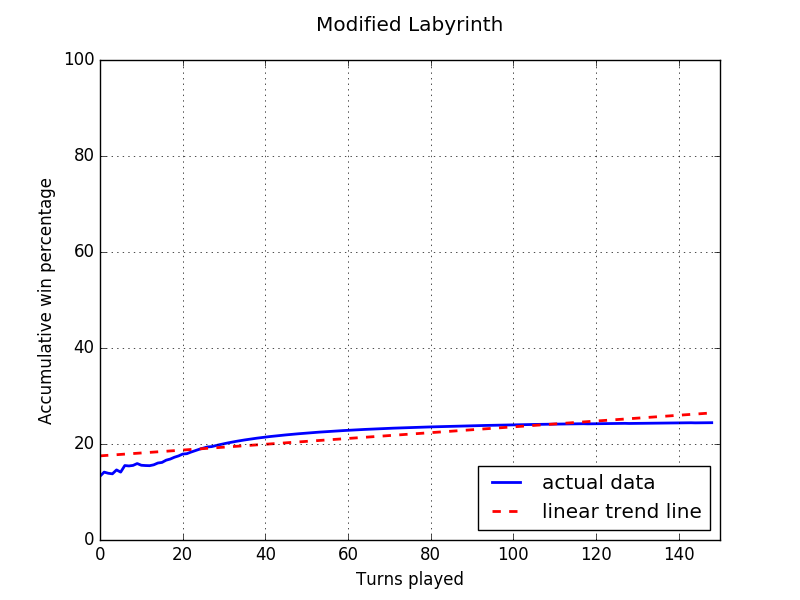}  
\caption{Modified Labyrinth average accumulative \%win
\label{fig:laby_score}
}
\end{figure}

The learning agent never won any single original Labyrinth game, but it did in the modified version. The cumulative winning percentage for this game, given in Figure \ref{fig:laby_score}, shows that the agent won more games in later turns, showing that our agent is capable of learning how to play Labyrinth with immediate rewards. It is worth to notice that, even with considerably low percentage ($15\%$), the MCTS agent won some original Labyrinth games. It might be beneficial using macro actions (repeatedly applying the same action in a sequence) for this type of games. In addition, the winning percentage of the MCTS agent increased significantly to $57\%$ with collectible items. This suggests that immediate reward signals could assist AI agents to solve games.

\subsubsection{Aliens}
\begin{figure}[!t]
\centering
\includegraphics[width=\textwidth/3]{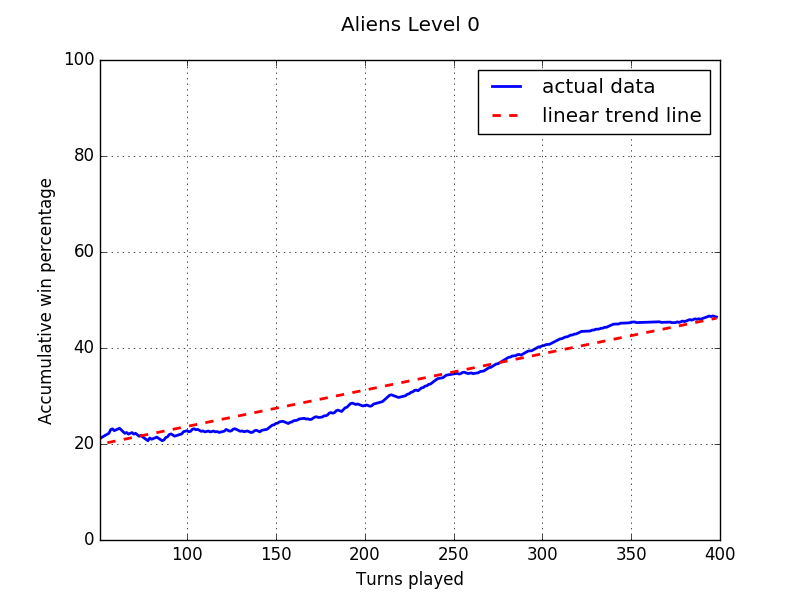}  
\caption{Aliens average accumulative win percentage
\label{fig:aliens_wp}
}
\end{figure}

Aliens is a stochastic shooting game in which the enemies move in one direction (above the agent). We measured the agent’s cumulative win percentage and plotted the results in Figure \ref{fig:aliens_wp}. The trend line shows that winning percentage increases with more episodes played, suggesting that the learning algorithm was also successful in this game. The MCTS agent outperformed our learning agent in this game with winning percentage $72\%$ and average score of approximately $77.5$. This might suggest that the planning algorithm suits stochastic games better than our learning method, in which the network might not be converged due to uncertainty in the game. 

\subsubsection{Sheriff}
\begin{figure}[!t]
\centering
\includegraphics[width=\textwidth/3]{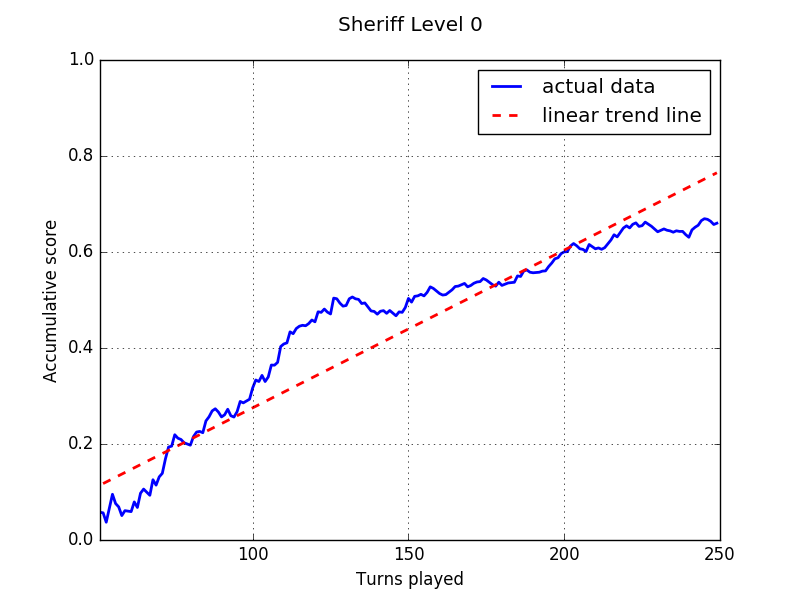}  
\caption{Sheriff average accumulative score}
\label{fig:sheriff_acc}
\end{figure}

Sheriff is another stochastic shooting game that is more complicated than Aliens. Specifically, enemies surround the player from all directions. The average cumulative score (since no victories were observed) was measured and presented in Figure \ref{fig:sheriff_acc}. The trend line of the data shows that the agent achieved a higher average score in later plays. That is, our agent was capable of learning to score more in this game. Similar to Aliens, the MCTS agent easily overcame our agent with a $94\%$ winning rate and almost perfect score on average ($7.96$ out of $8$). This confirms that uncertainty highly affected the learning agent performance more than the planning agent. Also, since the MCTS agent is able to guess the immediate future in crucial situations (such as almost being hit by the bullet), it was most likely to avoid it. This is in contrast to our agent, which required longer to actually 'learn' the connections between bullets in one and another position.

\subsubsection{Eggomania}
\begin{figure}[!t]
\centering
\includegraphics[width=\textwidth/3]{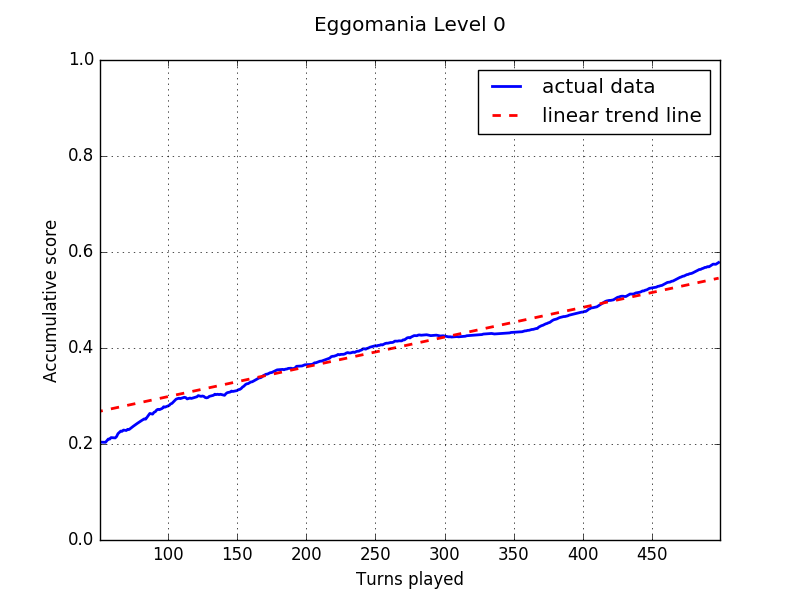}  
\caption{Eggomania average accumulative score}
\label{fig:eggo_acc}
\end{figure}

Eggomania is different to Aliens and Sheriff as the agent has to collect the objects dropped from the enemy instead of avoiding them. Failing to collect once will immediately cause it to lose the game. This is difficult since the agent might lose many times before an item is collected. Moreover, in order to win this game the agent must shoot the enemy after collecting items for a certain time, which is also challenging because the agent might incorrectly learn that the 'shoot' action was useless. Figure \ref{fig:eggo_acc} shows that the agent was gradually learning to achieve a higher score during training. Even though the MCTS agent suffered from this game nature (it won only 1 game out of 100), it managed to collect more items (about 5 in average) than our learning agent. 

\section{Conclusions and Future Work} \label{sec:conc}

In this paper, a screen capture learning agent for General Video Game Artificial Intelligence (GVG-AI) framework is presented for the first time. A Deep Q-Network algorithm was applied to develop such agent. Some improvements have been made to extend the original algorithm to work within the GVG-AI framework, such as supporting any screen size and non-visualise gameplay mode. The convolution neural network parameters were tuned as pre-experiment before being implemented as a part of the game agent in the real experiment. Results suggest that our learning agent was capable of learning to solve both static and stochastic games, as the accumulative wining percentage in static games and the accumulative average score in stochastic games increased with more games played. This suggests that the agent applied knowledge acquired during the earlier plays, to adapt to later repetitions in the same game. 
 
Several future works are possible in the agent's improvements. At the moment, the same CNN structure has been applied with every game experimented. However, it might be more efficient if the network can be scaled based on the game it is learning, since complicated games require larger networks. Another possible work involves transfer learning, as proposed by Braylan et. al. \cite{braylan2015reuse}, where an agent trained to play one game could apply the knowledge learned when playing other similar games faster than learning from scratch. There was an attempt to use this same idea in GVG-AI framework to improve the accuracy of the forward model \cite{braylan2016object}. This proves that the object knowledge in GVG-AI framework is transferable between games. This would be closer to the concept of human level intelligence or Artificial General Intelligence, as humans are capable of reusing their experiences from similar problems they have previously encountered. 

%



\ifCLASSOPTIONcaptionsoff
  \newpage
\fi



%
\bibliographystyle{ieeetr}
\bibliography{references}

\begin{thebibliography}{10}

\bibitem{lucas2007ms}
S.~M. Lucas, ``{Ms Pac-man Competition},'' {\em ACM SIGEVOlution}, vol.~2,
  no.~4, pp.~37--38, 2007.

\bibitem{togelius20102009}
J.~Togelius, S.~Karakovskiy, and R.~Baumgarten, ``{The 2009 Mario AI
  Competition},'' in {\em IEEE Congress on Evolutionary Computation}, pp.~1--8,
  IEEE, 2010.

\bibitem{goertzel2007artificial}
B.~Goertzel and C.~Pennachin, {\em {Artificial General Intelligence}}, vol.~2.
\newblock Springer, 2007.

\bibitem{perez20152014}
D.~Perez, S.~Samothrakis, J.~Togelius, T.~Schaul, S.~Lucas, A.~Cou{\"e}toux,
  J.~Lee, C.-U. Lim, and T.~Thompson, ``{The 2014 General Video Game Playing
  Competition},'' 2015.

\bibitem{levine2013general}
J.~Levine, C.~B. Congdon, M.~Ebner, G.~Kendall, S.~M. Lucas, R.~Miikkulainen,
  T.~Schaul, and T.~Thompson, ``{General Video Game Playing},'' {\em Dagstuhl
  Follow-Ups}, vol.~6, 2013.

\bibitem{schaul2013video}
T.~Schaul, ``{A Video Game Description Language for Model-based or Interactive
  Learning},'' in {\em Computational Intelligence in Games (CIG), 2013 IEEE
  Conference on}, pp.~1--8, IEEE, 2013.

\bibitem{competition_2007}
S.~M. Lucas, ``{Ms. Pac-man Competition: Screen Capture Version}.''
  http://cswww.essex.ac.uk/staff/sml/pacman/PacManContest.html, 2007.
\newblock {Accessed: 2016-12-31}.

\bibitem{kempka2016vizdoom}
M.~Kempka, M.~Wydmuch, G.~Runc, J.~Toczek, and W.~Ja{\'s}kowski, ``{ViZDoom: A
  Doom-based AI Research Platform for Visual Reinforcement Learning},'' {\em
  arXiv preprint arXiv:1605.02097}, 2016.

\bibitem{mnih2015human}
V.~Mnih, K.~Kavukcuoglu, D.~Silver, A.~A. Rusu, J.~Veness, M.~G. Bellemare,
  A.~Graves, M.~Riedmiller, A.~K. Fidjeland, G.~Ostrovski, {\em et~al.},
  ``{Human-level Control through Deep Reinforcement Learning},'' {\em Nature},
  vol.~518, no.~7540, pp.~529--533, 2015.

\bibitem{bellemare2012arcade}
M.~G. Bellemare, Y.~Naddaf, J.~Veness, and M.~Bowling, ``{The Arcade Learning
  Environment: An Evaluation Platform for General Agents},'' {\em Journal of
  Artificial Intelligence Research}, 2012.

\bibitem{genesereth2014general}
M.~Genesereth and M.~Thielscher, ``{General Game Playing},'' {\em Synthesis
  Lectures on Artificial Intelligence and Machine Learning}, vol.~8, no.~2,
  pp.~1--229, 2014.

\bibitem{browne2012survey}
C.~B. Browne, E.~Powley, D.~Whitehouse, S.~M. Lucas, P.~I. Cowling,
  P.~Rohlfshagen, S.~Tavener, D.~Perez, S.~Samothrakis, and S.~Colton, ``{A
  Survey of Monte Carlo Tree Search Methods},'' {\em IEEE Trans. on
  Computational Intelligence and AI in Games}, vol.~4, no.~1, pp.~1--43, 2012.

\bibitem{perez2014knowledge}
D.~Perez, S.~Samothrakis, and S.~Lucas, ``{Knowledge-based Fast Evolutionary
  MCTS for General Video Game Playing},'' in {\em 2014 IEEE Conference on
  Computational Intelligence and Games}, pp.~1--8, 2014.

\bibitem{perez2015open}
D.~Perez~Liebana, J.~Dieskau, M.~Hunermund, S.~Mostaghim, and S.~Lucas, ``{Open
  Loop Search for General Video Game Playing},'' in {\em Proceedings of the
  2015 Annual Conference on Genetic and Evolutionary Computation},
  pp.~337--344, ACM, 2015.

\bibitem{perez2016general}
D.~Perez-Liebana, S.~Samothrakis, J.~Togelius, S.~M. Lucas, and T.~Schaul,
  ``{General Video Game AI: Competition, Challenges and Opportunities},'' in
  {\em 30 AAAI Conference on Artificial Intelligence}, 2016.

\bibitem{samothrakis2015neuroevolution}
S.~Samothrakis, D.~Perez-Liebana, S.~M. Lucas, and M.~Fasli, ``{Neuroevolution
  for General Video Game Playing},'' in {\em 2015 IEEE Conference on
  Computational Intelligence and Games (CIG)}, pp.~200--207, 2015.

\bibitem{lecun2015deep}
Y.~LeCun, Y.~Bengio, and G.~Hinton, ``{Deep Learning},'' {\em Nature},
  vol.~521, no.~7553, pp.~436--444, 2015.

\bibitem{2_nielsen_2016}
M.~A. Nielsen, ``{Neural Networks and Deep Learning},'' 2016.

\bibitem{sutton1998reinforcement}
R.~S. Sutton and A.~G. Barto, {\em {Reinforcement Learning: An Introduction}},
  vol.~1.
\newblock MIT press Cambridge, 1998.

\bibitem{lange2010deep}
S.~Lange and M.~Riedmiller, ``{Deep Auto-encoder Neural Networks in
  Reinforcement Learning},'' in {\em The 2010 International Joint Conference on
  Neural Networks (IJCNN)}, pp.~1--8, IEEE, 2010.

\bibitem{ross2014general}
B.~Ross, {\em General Video Game Playing with Goal Orientation}.
\newblock PhD thesis, Master’s thesis, University of Strathclyde, 2014.

\bibitem{lecun1998gradient}
Y.~LeCun, L.~Bottou, Y.~Bengio, and P.~Haffner, ``Gradient-based learning
  applied to document recognition,'' {\em Proceedings of the IEEE}, vol.~86,
  no.~11, pp.~2278--2324, 1998.

\bibitem{krizhevsky2012imagenet}
A.~Krizhevsky, I.~Sutskever, and G.~E. Hinton, ``Imagenet classification with
  deep convolutional neural networks,'' in {\em Advances in neural information
  processing systems}, pp.~1097--1105, 2012.

\bibitem{mnih2016asynchronous}
V.~Mnih, A.~P. Badia, M.~Mirza, A.~Graves, T.~P. Lillicrap, T.~Harley,
  D.~Silver, and K.~Kavukcuoglu, ``{Asynchronous Methods for Deep Reinforcement
  Learning},'' {\em arXiv preprint arXiv:1602.01783}, 2016.

\bibitem{braylan2015reuse}
A.~Braylan, M.~Hollenbeck, E.~Meyerson, and R.~Miikkulainen, ``Reuse of neural
  modules for general video game playing,'' {\em arXiv preprint
  arXiv:1512.01537}, 2015.

\bibitem{braylan2016object}
A.~Braylan and R.~Miikkulainen, ``Object-model transfer in the general video
  game domain,'' in {\em Twelfth Artificial Intelligence and Interactive
  Digital Entertainment Conference}, 2016.

\end{thebibliography}

 


%








\end{document}